\documentclass[runningheads]{llncs}
\usepackage[T1]{fontenc}
\usepackage{tabularx}
\usepackage{graphicx,verbatim}
\usepackage{amsmath}
\usepackage{cite}
\usepackage{booktabs}
\usepackage{multirow}
\usepackage{siunitx} % For aligning decimals if needed
\usepackage{color, colortbl}
\usepackage[table]{xcolor}
\usepackage{moresize}
\usepackage{makecell}
\usepackage[colorlinks=true]{hyperref}
\usepackage{marvosym}
\hypersetup{
    linkcolor=red,  % 图的引用颜色为红色
    urlcolor=green, % 参考文献的颜色为绿色
}
\usepackage{tabularx}
\newcolumntype{Y}{>{\centering\arraybackslash}X}
\newcolumntype{L}{>{\raggedright\arraybackslash}X}

\newcommand{\ci}[2]{
    \makecell[tc]{
    #1\\[-1pt]
    {\textcolor[gray]{0.75}{#2}}%
    }
}

\begin{document}
\title{\texttt{Echo2ECG}: Enhancing ECG Representations with Cardiac Morphology from Multi-View Echos}

\author{
Michelle Espranita Liman\inst{1}\textsuperscript{(\Letter)}
\and
Özgün Turgut\inst{1}
\and \\
Alexander Müller\inst{2}
\and
Eimo Martens\inst{2}
\and \\
Daniel Rueckert\inst{1,3,4}
\and
Philip Müller\inst{1}
}
%index{Liman, Michelle Espranita}
%index{Turgut, Özgün}
%index{Müller, Alexander}
%index{Martens, Eimo}
%index{Rueckert, Daniel}
%index{Müller, Philip}

\authorrunning{M. E. Liman et al.}
\institute{
Chair for AI in Healthcare and Medicine, Technical University of Munich (TUM) and TUM University Hospital, Germany \\
\email{\{michelle.liman,oezguen.turgut,daniel.rueckert,philip.j.mueller\}@tum.de}
\and
Department of Cardiology, TUM University Hospital, Germany \\
\email{\{Alexander.Mueller,Eimo.Martens\}@mri.tum.de}
\and 
Department of Computing, Imperial College London, UK
\and
Munich Center for Machine Learning (MCML), Germany
}

\maketitle              % typeset the header of the contribution
\begin{abstract}

Electrocardiography (ECG) is a low-cost, widely used modality for diagnosing electrical abnormalities like atrial fibrillation by capturing the heart's electrical activity. However, it cannot directly measure cardiac morphological phenotypes, such as left ventricular ejection fraction (LVEF), which typically require echocardiography (Echo). Predicting these phenotypes from ECG would enable early, accessible health screening. Existing self-supervised methods suffer from a representational mismatch by aligning ECGs to \textit{single-view} Echos, which only capture local, spatially restricted anatomical snapshots. To address this, we propose \textbf{\texttt{Echo2ECG}}, a multimodal self-supervised learning framework that enriches ECG representations with the heart's morphological structure captured in \textit{multi-view} Echos. We evaluate \texttt{Echo2ECG} as an ECG feature extractor on two clinically relevant tasks that fundamentally require morphological information: (1) classification of structural cardiac phenotypes across three datasets, and (2) retrieval of Echo studies with similar morphological characteristics using ECG queries. Our extracted ECG representations consistently outperform those of state-of-the-art unimodal and multimodal baselines across both tasks, despite being 18$\times$ smaller than the largest baseline. These results demonstrate that \texttt{Echo2ECG} is a robust, powerful ECG feature extractor. Our code is accessible at: \href{https://github.com/michelleespranita/Echo2ECG}{https://github.com/michelleespranita/Echo2ECG}.


\end{abstract}
\section{Introduction}

Electrocardiography (ECG) is a cost-effective and ubiquitous tool for cardiac assessment. While it is highly effective at diagnosing electrical heart abnormalities, such as atrial fibrillation \cite{doi:10.1161/CIRCEP.125.013734}, it cannot directly quantify cardiac morphological phenotypes like left ventricular ejection fraction (LVEF). Traditionally, these structural measurements require echocardiography (Echo), a modality often limited by cost, medical expertise, and accessibility. Recently, AI-based ECG models have demonstrated the ability to detect morphological phenotypes such as left-sided valvular heart disease \cite{doi:10.1016/j.jacc.2022.05.029} and hypertrophic cardiomyopathy \cite{doi:10.1016/j.jacc.2019.12.030}, enabling early and accessible health screening.

Current methods for transferring morphological information from Echo to ECG primarily rely on supervised training using Echo-derived labels \cite{doi:10.1016/j.jacc.2022.05.029, 10.1093/eurheartj/ehab153, doi:10.1016/j.jacc.2019.12.030, Poterucha2025}. However, they fail to leverage large-scale unlabeled data through self-supervised learning, which is critical for building robust ECG representations that could be used for other prediction tasks.

Recent multimodal self-supervised methods, most notably EchoingECG \cite{GaoYua_EchoingECG_MICCAI2025}, attempt to bridge this gap by pre-training on ECG, Echo, and text-based discharge summaries. By aligning ECGs to \textit{single-view} Echos (e.g., apical four-chamber/A4C), the approach attempts to map global electrical signals to local, spatially restricted anatomical snapshots. This creates a representational mismatch: a single-view Echo captures only a subset of the heart’s structure, whereas the ECG reflects the heart’s global electrical activity. Furthermore, by requiring paired text reports, the model’s scalability is restricted to datasets where clinical reports are available.

\begin{sloppypar}
To address these limitations, we propose \texttt{Echo2ECG}, a multimodal self-supervised learning framework that enriches ECG representations with morphological information from \textit{multi-view} Echo studies. By leveraging powerful unimodal encoders pre-trained on large datasets, our method enables distillation of a complete picture of the heart's morphological structure into ECG representations without depending on paired text data. The resulting ECG encoder serves as a powerful ECG feature extractor for downstream tasks requiring morphological information. Our contributions are as follows:
\end{sloppypar}

\begin{itemize}
    \item We propose \texttt{\textbf{Echo2ECG}}, a multimodal self-supervised learning framework that aligns global ECGs to multi-view Echo studies to learn rich, structure-aware ECG representations.
    \item We evaluate \texttt{Echo2ECG} as an ECG feature extractor for the classification of structural cardiac phenotypes across three datasets. Our extracted ECG features consistently outperform those of state-of-the-art unimodal and multimodal baselines, despite our model being 18$\times$ smaller than the largest baseline. This demonstrates that morphological information can be effectively distilled into a lightweight yet powerful ECG feature extractor.
    \item We demonstrate the alignment of our learned representation space by successfully retrieving Echo studies with similar morphological characteristics using only ECG queries.
\end{itemize}

\section{Related Work}

The increasing accessibility of large-scale multimodal datasets, such as MIMIC-IV \cite{PhysioNet-mimiciv-3.1} and the UK Biobank \cite{Sudlow2015-jd}, has enabled research into cross-modal knowledge transfer in the medical domain. Most multimodal learning frameworks are inspired by CLIP \cite{pmlr-v139-radford21a}, which aligns modalities by pulling paired representations together and pushing unpaired ones apart in a joint embedding space. In the cardiac domain, MMCL \cite{TURGUT2025103451} and PTACL \cite{SelAle_Global_MICCAI2025} use this approach to distill cardiac morphological information from Cardiac MRI (CMR) into ECG. Most similar to our work, EchoingECG \cite{GaoYua_EchoingECG_MICCAI2025} aligns ECGs with single-view Echos and discharge reports. These studies consistently demonstrate that aligning ECGs with "gold standard" imaging modalities significantly improves the prediction of structural cardiac phenotypes solely based on ECG.

Concurrently, large-scale ECG foundation models such as xECG \cite{lunelli2025benchecgxecgbenchmarkbaseline} and ECGFounder \cite{li2025electrocardiogram}, and time-series foundation models such as OTIS \cite{turgut2025generalisabletimeseriesunderstanding}, have emerged, utilizing self-supervised learning on massive ECG datasets like MIMIC-IV-ECG \cite{PhysioNet-mimic-iv-ecg-1.0} and CODE \cite{Ribeiro2019-aa}. We argue that while these unimodal models are powerful, their performance on structural cardiac phenotype prediction can be further enhanced through multimodal alignment. To this end, we initialize the ECG encoder in our framework with OTIS \cite{turgut2025generalisabletimeseriesunderstanding}.

Our work diverges from existing literature in two key ways. First, we align ECGs with complete multi-view Echo studies, solving the representational mismatch that occurs when aligning ECGs with individual Echo views. Secondly, unlike EchoingECG, we focus strictly on ECG-Echo relationship by omitting the usage of text reports during pre-training. To our knowledge, this is the first study to investigate the direct alignment of ECGs with full multi-view Echo studies.

\section{Method}

\begin{figure}[t!]
    \centering
        \includegraphics[width=\linewidth]{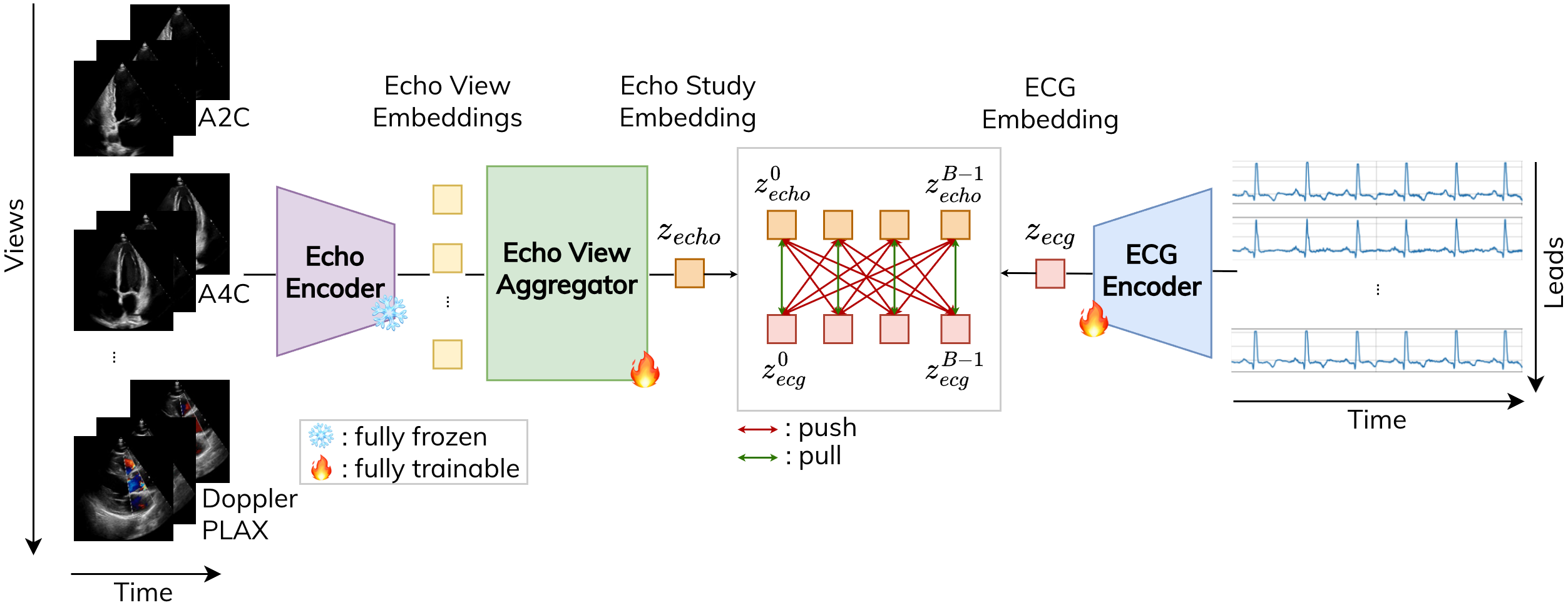}
    \caption{
    Overview of \texttt{\textbf{Echo2ECG}}.
    To address the representational mismatch in prior methods aligning ECGs with \textit{single-view} Echos, our framework uses contrastive learning to align ECGs with \textit{multi-view} Echo studies. This multi-view alignment distills comprehensive cardiac morphological information from Echos into ECG representations.
    We aggregate view-level Echo embeddings from a frozen, powerful Echo encoder into a study-level Echo embedding. During pre-training, we optimize only the ECG encoder, the Echo view aggregator, and modality-specific projection layers. The ECG encoder is then used to extract ECG representations for subsequent downstream tasks.
    }
    \label{fig:echo2ecg_pipeline}
\end{figure}

The goal of our proposed method is to transfer rich cardiac morphological information from echocardiography into ECG representations. This enables the prediction of Echo-derived structural cardiac phenotypes, such as left ventricular ejection fraction (LVEF), using the widely accessible and low-cost ECG modality alone. As shown in Figure \ref{fig:echo2ecg_pipeline}, our framework comprises four main components: (1) an ECG encoder that maps an ECG signal to an embedding; (2) an Echo encoder that maps each Echo view to an embedding; (3) an Echo view aggregation module that combines multiple view-level embeddings from the same study into a single study-level representation; and (4) modality-specific projection layers.

\subsection{ECG and Echo Encoders}

We initialize the ECG and Echo encoders using strong, pre-trained unimodal models to leverage large-scale prior knowledge. The ECG encoder is a 12-layer tiny transformer \cite{2017_Vaswani_Transformer}, initialized from OTIS \cite{turgut2025generalisabletimeseriesunderstanding}, which was pre-trained on a multi-domain time-series corpus including MIMIC-IV-ECG \cite{PhysioNet-mimic-iv-ecg-1.0}. Each ECG is represented by the mean token embedding. The Echo encoder is a MViTv2 \cite{2022_Li_mvitv2}, initialized from EchoPrime \cite{Vukadinovic2025}, which was trained via multimodal contrastive learning between Echo views and clinical reports from an in-house Cedars-Sinai dataset. Each Echo view is represented by the [CLS] token embedding.

The ECG encoder, the Echo view aggregator, and the projection layers are fully trainable, while the Echo encoder is kept frozen. Overall, the framework has 12.5M trainable parameters and is trained for 50 epochs. Hyperparameter details are available in the code.

\subsection{Echo View Aggregation}

A single Echo view captures only part of the heart, e.g., the A4C view captures the four chambers of the heart but not the aortic valve, which is visible in the PSAX view. In clinical practice, cardiologists do not base diagnoses on a single view; instead, they integrate information across multiple views from a complete Echo study to obtain a comprehensive cardiac assessment. Similarly, by distilling information from the full Echo study rather than isolated views into an ECG, we ensure that the resulting ECG features are informed by the entire heart.

To achieve this, each Echo view is first encoded independently by the Echo encoder to produce view-level embeddings. These embeddings are projected from 512- to 1024-dim and aggregated via attention pooling into a single study-level embedding. This allows the model to weigh views by relevance while preserving information from the entire study.

\subsection{Multimodal Contrastive Pre-Training}

To transfer information from multi-view Echo studies to ECGs, we align the embeddings from both modalities within a shared 512-dim latent space using the CLIP\cite{pmlr-v139-radford21a} contrastive objective. After modality-specific projection layers, the ECG and Echo study embeddings are pulled together for matched pairs and pushed apart for unmatched pairs within a batch.

For a batch of size $B$, the ECG$\rightarrow$Echo contrastive loss is defined as:

\begin{equation}
    \mathcal{L}_{\text{ECG}\rightarrow\text{Echo}}
    = -\frac{1}{B} \sum_{i=1}^{B}
    \log \frac{\text{exp}(\text{cos}(z_{\text{ecg}}^{i}, z_{\text{echo}}^{i})/\tau)}{\sum_{j=1}^{B} \text{exp}(\text{cos}(z_{\text{ecg}}^{i}, z_{\text{echo}}^{j})/\tau)}
\end{equation}

where $\tau$ denotes the temperature parameter, and $z_{\text{ecg}}$ and $z_{\text{echo}}$ are the ECG and Echo study embeddings after the projection layers respectively. An analogous Echo$\rightarrow$ECG loss is computed by swapping the modalities. The final training objective is a weighted combination of the two directional losses.

\section{Experimental Setup}

\subsection{Dataset and Pre-processing}

\begin{sloppypar}
We pre-train using paired ECG-Echo studies collected at our institution, TUM Klinikum Rechts der Isar. An ECG and an Echo study of a patient are a valid pair if they were acquired within 7 days. We split the data at the patient level (Table \ref{tab:data_splits}).  
\end{sloppypar}

\begin{table}[t]
    \centering
    \fontsize{8}{9.5}\selectfont
    \caption{Internal pre-training dataset details from TUM Klinikum Rechts der Isar. Age is presented as Mean $\pm$ SD.}
    \label{tab:data_splits}
    \setlength{\tabcolsep}{1pt}
    \begin{tabularx}{\textwidth}{l YYYYYYY}
        \toprule
        \textbf{Split} & \textbf{\#Pairs} & \textbf{\#Patients} & \textbf{\#ECGs} & \textbf{\makecell[c]{\#Echo\\Studies}} & \textbf{\makecell[c]{\#Echo\\Views}} & \textbf{Age} & \textbf{\makecell[c]{Sex\\(\%Male)}} \\
        \midrule
        Train & 11,859 & 6,836 & 11,516 & 8,712 & 179,571 & 65.6$\pm$16.1 & 63.1\% \\
        Val   & 1,248  & 565   & 1,213  & 934 & 20,633 & 65.3$\pm$16.7 & 64.7\%  \\
        Test  & 1,358 & 566   & 1,290  & 924 & 20,903 & 64.8$\pm$16.6 & 64.1\%  \\
        \bottomrule
    \end{tabularx}
\end{table}

ECGs (12-lead, 10-second, 500 Hz) are processed following OTIS \cite{turgut2025generalisabletimeseriesunderstanding}, including baseline drift removal using asymmetric least-squares smoothing and lead-group normalization. We use $T=1008$ timesteps and apply the same augmentations as in OTIS.

Echo studies include up to 128 views. Following EchoPrime \cite{Vukadinovic2025}, we extract the Echo regions, remove annotations, and downsample the spatial resolution to 224. We sample 16 frames with stride 2, and zero-pad shorter clips. Each Echo view is normalized using EchoPrime train set statistics without augmentation.

\subsection{Baselines}

We compare our ECG model against two categories of baselines: (1) unimodal ECG models, such as OTIS (7.1M params) \cite{turgut2025generalisabletimeseriesunderstanding}, xECG (57.0M) \cite{lunelli2025benchecgxecgbenchmarkbaseline}, and ECGFounder (30.7M) \cite{li2025electrocardiogram}; and (2) multimodal ECG models, such as EchoingECG (126.6M) \cite{GaoYua_EchoingECG_MICCAI2025} and PTACL (19.3M) \cite{SelAle_Global_MICCAI2025}. To ensure fair comparison, we follow the data pre-processing and training protocols in the original papers or codebases.

\subsection{Evaluation Strategy}

\subsubsection{Structural Cardiac Phenotype Prediction}

We assess the quality of the extracted ECG representations on two downstream tasks which require morphological information: (1) left ventricular ejection fraction (LVEF) classification (reduced $\le40\%$, mildly reduced $40-50\%$, and normal $\ge50\%$ \cite{10.1093/eurheartj/ehab368}) and (2) structural heart disease (SHD) classification (absence vs. presence of moderate/greater disease \cite{PhysioNet-echonext-1.1.0}). For both tasks, we train a $k$-nearest neighbors classifier (kNN) with $k=5$ on the ECG features. As kNN performance is deterministic, we report AUROC with 95\% CI via 1,000 bootstrap resamples of the test set.

Downstream dataset details are summarized in Table \ref{tab:downstream_dataset}. To assess generalization, models are evaluated across three cohorts on LVEF classification: our internal dataset, EchoNext \cite{PhysioNet-echonext-1.1.0}, and UK Biobank  (UKB) \cite{Sudlow2015-jd}. For SHD classification on EchoNext, we utilize the provided splits and evaluate the strength of the ECG embeddings by scaling training data from 0.1\% to 100\%.

\begin{table}[t!]
    \centering
    \fontsize{8}{9.5}\selectfont
    \caption{Downstream datasets and label distributions. LVEF: 0/1/2 = reduced/mildly reduced/normal; SHD: 0/1 = absence/presence. These tasks assess how well the ECG representations capture morphological information.}
    \label{tab:downstream_dataset}
    \begin{tabularx}{\textwidth}{p{2.5cm}p{1.7cm} L L L}
        \toprule
        \textbf{Task} & \textbf{Dataset} & \textbf{Train ($N$)} & \textbf{Val ($N$)} & \textbf{Test ($N$)} \\
        \midrule
        \multirow{3}{*}{LVEF clf (0/1/2)}
        & Internal
        & 294 (33/33/33\%)
        & 564 (11/16/73\%)
        & 562 (14/15/71\%) \\
        & EchoNext
        & 300 (33/33/33\%)
        & 4,112 (17/7/76\%)
        & 4,827 (16/7/77\%) \\
        & UKB
        & 255 (33/33/33\%)
        & 2,834 (1/4/95\%)
        & 2,827 (1/4/95\%) \\
        \midrule
        SHD clf (0/1)
        & EchoNext
        & 72,475 (48/52\%)
        & 4,626 (57/43\%)
        & 5,442 (57/43\%) \\
        
        \bottomrule
    \end{tabularx}
\end{table}

\subsubsection{Retrieval of Echo Studies with Similar Phenotypes using ECG Queries}

We assess the learned joint embedding space via phenotype-aware cross-modal retrieval. Given a query ECG, Echo study embeddings are ranked by cosine similarity, and the top-$k$ Echo studies are retrieved. A retrieved Echo study is a match if its phenotype value lies within $\pm0.5\sigma$ of the query ECG’s phenotype, where $\sigma$ denotes the standard deviation of that phenotype in the test set. For models without an Echo view aggregator, the Echo study embedding is calculated by averaging the embeddings of the Echo views belonging to the study. We evaluate four cardiac phenotypes: LV end-diastolic and end-systolic volumes (EDV, ESV), stroke volume (SV), and ejection fraction (EF); and report Precision@1 (i.e., $k=1$) and mean rank (MnR) across the test set.

\section{Results and Discussion}

\subsection{\texttt{Echo2ECG} Extracts Powerful Out-of-the-Box Features}

\begin{table*}[t!]
    \centering
    \fontsize{8}{9.5}\selectfont
    \caption{
    \textbf{Best} and \underline{second-best} AUROC [95\% CI] ($\uparrow$) for ECG-based LVEF classification.
    \texttt{Echo2ECG} provides high-quality, out-of-the box ECG features, achieving the highest AUROC on our internal dataset and on EchoNext. On UKB, it ranks second behind PTACL, which was pre-trained on UKB by pairing ECG with CMR—a superior modality to Echo for LVEF quantification.
    }
    \label{tab:lvef_clf}
    \begin{tabularx}{\textwidth}{p{1.7cm}lp{1.2cm} *{3}{Y}}
        \toprule

        \multirow{2}{*}{\textbf{\makecell[l]{Paired\\Modality}}} & \multirow{2}{*}{\textbf{Method}} & & \textbf{Internal} & \multicolumn{2}{c}{\textbf{External}} \\
        \cmidrule(lr){4-4} \cmidrule(lr){5-6}
        & & & \textbf{Internal} & \textbf{EchoNext} & \textbf{UKB} \\
        
        \midrule
        
        - & OTIS
            & (7.1M)
            & \ci{0.688}{[0.650, 0.727]}
            & \ci{0.697}{[0.680, 0.713]}
            & \ci{0.637}{[0.574, 0.694]} \\
        - & xECG
            & (57.0M)
            & \ci{0.732}{[0.696, 0.769]}
            & \ci{0.695}{[0.677, 0.711]}
            & \ci{0.531}{[0.474, 0.591]} \\
        - & ECGFounder
            & (30.7M)
            & \ci{\underline{0.746}}{[0.710, 0.781]}
            & \ci{0.686}{[0.670, 0.702]}
            & \ci{0.653}{[0.592, 0.708]} \\
        Text+Echo & EchoingECG
            & (126.6M)
            & \ci{0.740}{[0.703, 0.776]}
            & \ci{0.676}{[0.661, 0.692]}
            & \ci{0.604}{[0.541, 0.662]} \\
        CMR & PTACL
            & (19.3M)
            & \ci{0.723}{[0.686, 0.759]}
            & \ci{\underline{0.704}}{[0.689, 0.719]}
            & \ci{\textbf{0.729}}{[0.677, 0.770]} \\
        \rowcolor{blue!5}
        \textbf{Echo} & \texttt{\textbf{Echo2ECG}}
            & (7.1M)
            & \ci{\textbf{0.785}}{[0.750, 0.819]}
            & \ci{\textbf{0.723}}{[0.708, 0.738]}
            & \ci{\underline{0.692}}{[0.642, 0.736]} \\
        \bottomrule
    \end{tabularx}
\end{table*}

\begin{figure}[t]
    \centering
        \includegraphics[width=0.9\linewidth]{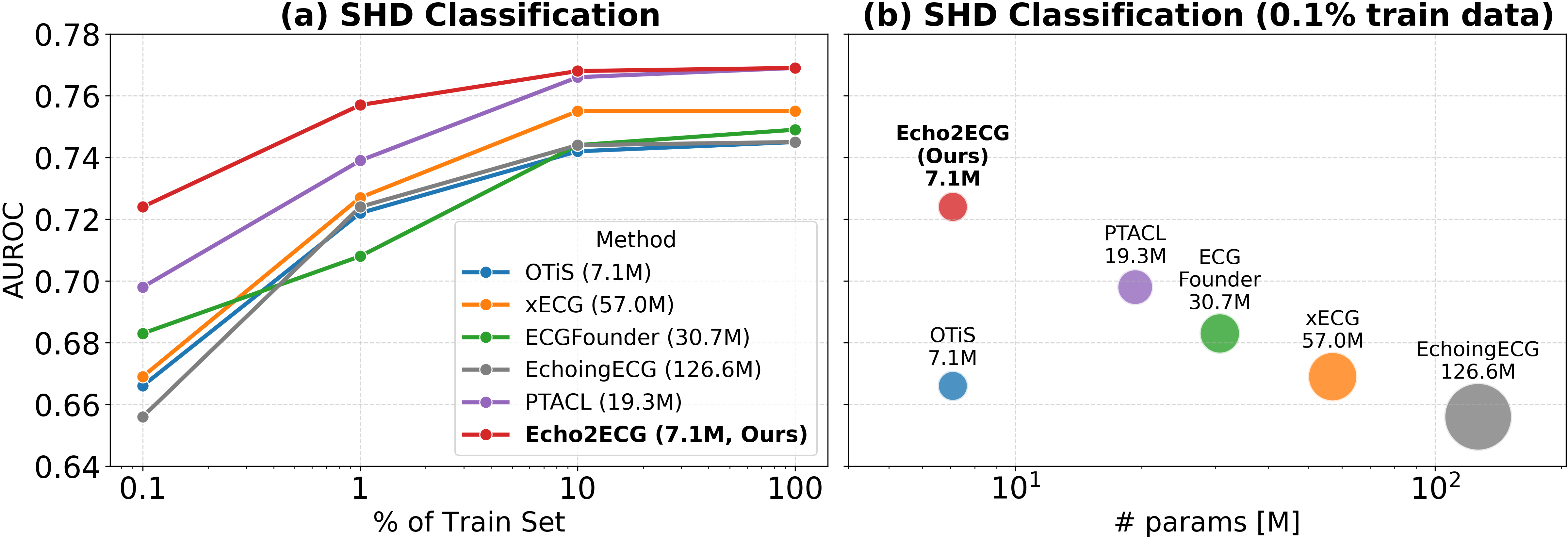}
    \caption{
    AUROC ($\uparrow$) for ECG-based SHD classification.
    (a) A kNN classifier trained on just 1\% of data using ECG representations extracted by \texttt{Echo2ECG} outperforms most models trained on 100\% of the data, meaning that \texttt{Echo2ECG} provides robust ECG features that work well under low-data regimes.
    (b) \texttt{Echo2ECG} outperforms EchoingECG at $0.1\%$ training data, despite being 18$\times$ smaller.
    }
    \label{fig:shd_clf}
\end{figure}

To evaluate diagnostic utility, we assess \texttt{Echo2ECG} on (i) LVEF (Table \ref{tab:lvef_clf}) and (ii) SHD classification (Figure \ref{fig:shd_clf}) against unimodal and multimodal baselines. Across both tasks, \texttt{Echo2ECG} consistently achieves superior performance.

For LVEF classification, a kNN classifier trained on \texttt{Echo2ECG}'s ECG features achieves the highest AUROC on our internal dataset and on EchoNext, improving AUROC by 5.2\% and 2.7\% respectively over the second-best model on each dataset. On UKB, our model ranks second behind PTACL, which was pre-trained on UKB by pairing ECG with CMR—a superior modality to Echo for LVEF quantification.

For SHD classification, a kNN classifier trained on just 1\% of the ECG features extracted by \texttt{Echo2ECG} outperforms nearly all other models trained on the full 100\% dataset (Figure \ref{fig:shd_clf}(a)), showing that \texttt{Echo2ECG} provides robust ECG features that work well under low-data regimes. Remarkably, a kNN trained on just 0.1\% of the ECG features extracted by \texttt{Echo2ECG} outperforms all competing models trained on the same amount of data—including EchoingECG—while \texttt{Echo2ECG} remains 18$\times$ smaller (Figure \ref{fig:shd_clf}(b)).

\subsection{Multi-View Alignment Enables Accurate Morphological Cross-Modal Retrieval}

In phenotype-aware ECG-to-Echo retrieval (Table \ref{tab:retrieval}), \texttt{Echo2ECG} achieves the highest Prec@1 and lowest MnR. Our study-level, multi-view alignment outperforms single-view alignment, highlighting the importance of aligning global electrical signals with comprehensive anatomy to encode morphological structure. Interestingly, EchoingECG performs worse than random retrieval, likely because its contrastive objective prioritizes ECG-Text over ECG-Echo alignment \cite{GaoYua_EchoingECG_MICCAI2025}, failing to directly link the heart's electrical activity with its morphological structure.

\begin{table*}[t!]
    \centering
    \fontsize{8}{9.5}\selectfont
    \caption{Cross-modal retrieval performance for structural cardiac phenotypes. Multi-view alignment allows \texttt{Echo2ECG} to accurately retrieve Echo studies from ECG queries.}
    \label{tab:retrieval}
    \begin{tabularx}{\textwidth}{l *{8}{Y}}
        \toprule
        \multirow{2}{*}{\textbf{Method}} & \multicolumn{2}{c}{\textbf{LVEDV}} & \multicolumn{2}{c}{\textbf{LVESV}} & \multicolumn{2}{c}{\textbf{LVSV}} & \multicolumn{2}{c}{\textbf{LVEF}} \\
        \cmidrule(lr){2-3} \cmidrule(lr){4-5} \cmidrule(lr){6-7} \cmidrule(lr){8-9}
        & P@1$\uparrow$ & MnR$\downarrow$ & P@1$\uparrow$ & MnR$\downarrow$ & P@1$\uparrow$ & MnR$\downarrow$ & P@1$\uparrow$ & MnR$\downarrow$ \\
        \midrule
        Random retrieval
        & 0.296 & 5.915
        & \underline{0.366} & 6.081
        & 0.323 & \underline{4.570}
        & 0.295 & 7.230 \\
        EchoingECG
        & 0.241 & 10.111
        & 0.337 & 10.998
        & 0.325 & 5.320
        & 0.232 & 13.035 \\
        \texttt{Echo2ECG} (\textit{single}-view Echo)
        & \underline{0.332} & \underline{5.376}
        & \textbf{0.517} & \underline{4.366}
        & \underline{0.334} & 5.125
        & \textbf{0.415} & \underline{5.346} \\
        \rowcolor{blue!5}
        \texttt{Echo2ECG} (\textit{multi}-view Echo)
        & \textbf{0.387} & \textbf{4.358}
        & \textbf{0.517} & \textbf{3.841}
        & \textbf{0.379} & \textbf{4.193}
        & \underline{0.404} & \textbf{5.101} \\
        \bottomrule
    \end{tabularx}
\end{table*}

\subsection{Learnable Pooling of Multi-View Echo Benefits Alignment}

Our ablation study (Table \ref{tab:ablation}) shows that multi-view alignment consistently outperforms single-view alignment, with attention pooling proving most effective at distilling key morphological features into ECG representations.

\begin{table*}[t!]
    \centering
    \fontsize{8}{9.5}\selectfont
    \caption{Ablation of \texttt{Echo2ECG} components. Mean AUROC [95\% CI] ($\uparrow$) is reported. Multi-view alignment and attention pooling improve the quality of ECG features.}
    \label{tab:ablation}
    \setlength{\tabcolsep}{5pt}
    \begin{tabularx}{\textwidth}{ll *{2}{Y}}
        \toprule
        \textbf{Alignment} & \textbf{View Agg.} & \textbf{\makecell[c]{LVEF clf\\(Internal)}} & \textbf{\makecell[c]{SHD clf\\(EchoNext 1\%)}} \\
        \midrule
        ECG $\leftrightarrow$ \textit{single}-view Echo
        & -
        & 0.747
        & 0.743 \\
        ECG $\leftrightarrow$ \textit{multi}-view Echo
        & mean pool
        & \underline{0.775}
        & 0.752 \\
        ECG $\leftrightarrow$ \textit{multi}-view Echo
        & CLS pool
        & 0.756
        & \underline{0.754} \\
        \rowcolor{blue!5}
        ECG $\leftrightarrow$ \textit{multi}-view Echo
        & attention pool
        & \textbf{0.785}
        & \textbf{0.757} \\
        \bottomrule
    \end{tabularx}
\end{table*}

\section{Conclusion}

We present \texttt{Echo2ECG}, a self-supervised framework that enriches ECG representations with cardiac morphology from multi-view Echo studies. Particularly, we resolve the representational mismatch of prior single-view methods and eliminate reliance on paired text data. Across three datasets, our lightweight ECG model serves as a powerful feature extractor, outperforming state-of-the-art baselines in structural phenotyping and morphology-aware cross-modal Echo retrieval. Remarkably, it achieves this while being 18$\times$ smaller than the largest baseline.

Our work has several limitations. First, we do not account for beat- or phase-level synchronization between ECGs and Echos, which may limit the model’s ability to capture temporal relationships between electrical activity in ECGs and mechanical activity in Echos. Second, although we do not rely on text reports, our method requires paired ECG–Echo studies, which may be scarce at scale.

\begin{credits}
\subsubsection{\ackname}
This study was approved by the Ethics Committee of TUM Klinikum Rechts der Isar (reference number 2025-395-S-CB, application dated July 13, 2025). This research has been conducted using the UK Biobank Resource under Application Number 87802.

\subsubsection{\discintname}
The authors have no competing interests to declare that
are relevant to the content of this article.
\end{credits}

\bibliographystyle{splncs04}
\bibliography{Paper-1932}

\end{document}